%% file: main.tex
\documentclass[11pt]{article}
\usepackage{amsmath,amssymb,graphicx,booktabs,float,wrapfig}
\usepackage{array,tabularx,etoolbox}
\usepackage[section]{placeins}
\usepackage{ziplab-tech-report}
\usepackage{hyperref,xurl}

\renewcommand{\techreportfield}[2]{%
  \ifdefempty{#2}{}{\noindent{\ziplabmetalabel{#1}#2\par}}}
\AtBeginEnvironment{table}{%
  \setlength{\belowcaptionskip}{5pt}%
  \renewcommand{\arraystretch}{1.12}}
\definecolor{deepblue}{RGB}{36,82,128}
\definecolor{deepred}{RGB}{166,62,55}
\definecolor{deepgreen}{RGB}{51,120,91}
\definecolor{softgray}{RGB}{242,244,247}

\newcommand{\reporttitle}{MixTok: Adaptive Visual Tokenization\\
with Global 1D Tokens and Sparse 2D Refinement}
\newcommand{\reportauthors}{%
Chenge Li$^{1*}$\quad Yefei He$^{1*\dagger}$\quad Xinyu Liu$^{2}$\\[0.25em]
Feng Chen$^{3\dagger}$\quad Hong Zhou$^{1}$\quad Bohan Zhuang$^{1}$}
\newcommand{\reportaffiliations}{%
$^{1}$Zhejiang University\quad
$^{2}$The Hong Kong University of Science and Technology\\[0.15em]
$^{3}$University of Adelaide}
\techreportlabel{ZIP Lab Technical Report}
\techreportshorttitle{MixTok}
\techreportdate{}
\techreportconference{}
\techreportorcid{}
\techreportproject{\url{https://lcglichenge.github.io/MixTok-page/}}
\techreportcode{\url{https://github.com/LCGLiChenge/MixTok}}
\techreportemail{\href{mailto:billhe@zju.edu.cn}{billhe@zju.edu.cn};
\href{mailto:chenfeng1271@gmail.com}{chenfeng1271@gmail.com}}
\techreportkeywords{Visual tokenization, adaptive token allocation, image generation}
\techreportfootnote{* Equal contribution. $^{\dagger}$Corresponding author.}
\techreportlogos{\techreportlogo{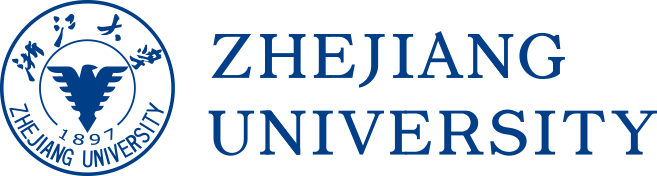}\hspace{0.4em}\techreportlogo{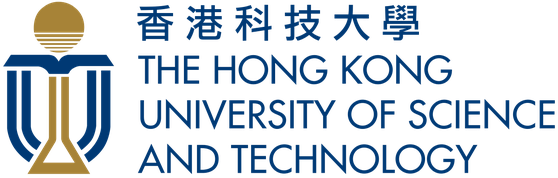}\hspace{0.4em}\techreportlogo{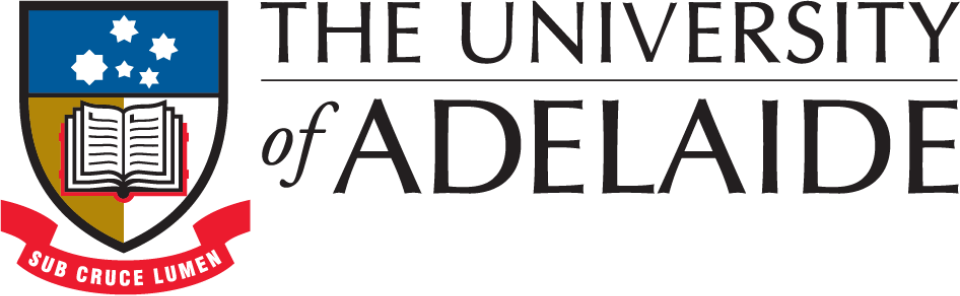}}
\hypersetup{
  pdftitle={MixTok: Adaptive Visual Tokenization with Global 1D Tokens and Sparse 2D Refinement},
  pdfauthor={Chenge Li, Yefei He, Xinyu Liu, Feng Chen, Hong Zhou, Bohan Zhuang},
  colorlinks=true, linkcolor=ziplabblue, citecolor=ziplabblue, urlcolor=ziplabblue
}

\begin{document}
\raggedbottom
\ziplabmaketitle{\reporttitle}{\reportauthors}{\reportaffiliations}{%
\begin{center}
\begin{minipage}{\textwidth}

    \centering
    \includegraphics[width=\linewidth]{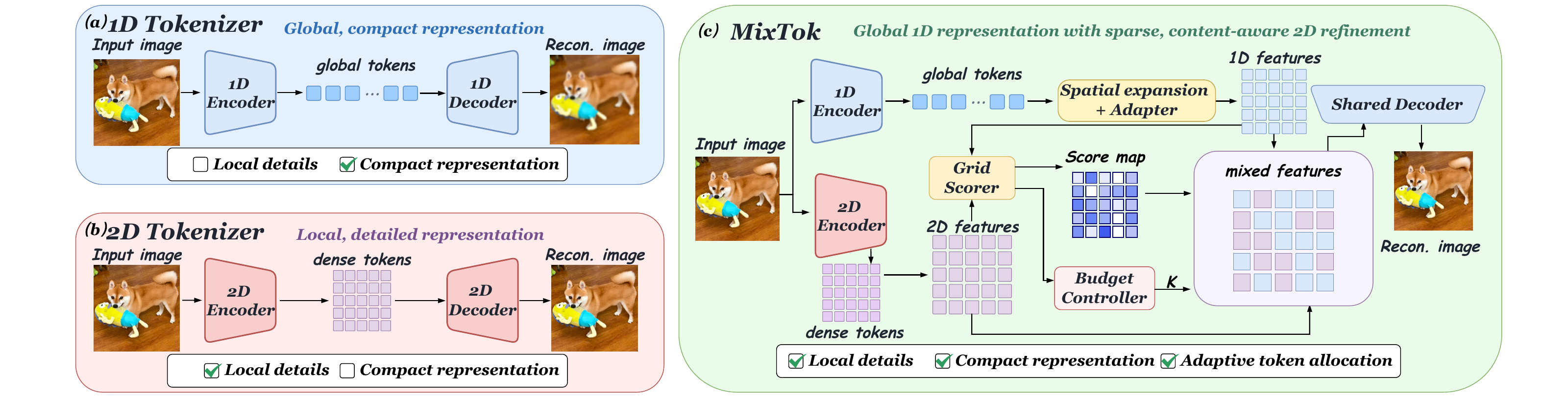}
    \captionsetup{hypcap=false}
\captionof{figure}{\textbf{Comparison of different visual tokenization paradigms.} (a) 1D tokenizer provides a compact global representation but may lose fine-grained local details. (b)  2D tokenizer preserves local details using a dense spatial token grid, but incurs substantial token overhead. (c) Our MixTok combines compact 1D tokens with content-adaptively selected 2D tokens, achieving both representation compactness and local-detail preservation.}
    \label{fig:MixTok-comparison}

\end{minipage}
\end{center}
}
\ziplabendtitle{\input{sections/abstract}}
\clearpage

\section{Introduction}
\label{sec:introduction}
\input{sections/introduction_yefei}

\section{Related Work}
\label{sec:related-work}
\input{sections/relatedwork_yefei}

\section{Preliminary}
\label{sec:preliminary}
\input{sections/preliminaries_yefei}

\section{Method}
\label{sec:method}
\suppressfloats[t]
\input{sections/42_43_lichenge}

\section{Experiment}
\label{sec:experiment}
\suppressfloats[t]
\input{sections/experiment}

\section{Conclusion and Future Work}
\label{sec:conclusion-and-future-work}
\input{sections/conclusion_and_future_work}

\par\medskip
\subsection*{AI use statement}
\input{sections/ai_use}

\subsection*{Ethics statement}
\input{sections/ethics_statement}

\subsection*{Reproducibility statement}
\input{sections/reproducibility_statement}

\begingroup
\small
\setlength{\bibsep}{0pt}
\bibliography{references}
\bibliographystyle{unsrtnat}
\endgroup

\clearpage
\appendix
\section{Appendix}
\label{app:appendix}
\input{sections/appendix}
\end{document}

%% file: sections/abstract.tex
Efficient visual tokenization requires balancing representation compactness with the preservation of fine-grained image content. Conventional grid-based tokenizers allocate tokens uniformly across spatial locations despite non-uniform visual complexity, whereas compact 1D tokenizers aggregate global information but can struggle to preserve local details. We introduce MixTok, a hybrid visual tokenizer that combines a compact 1D global representation with sparse 2D tokens for local refinement. A content-adaptive router ranks candidate locations by refinement priority and determines the number of 2D tokens for each image under a target average token budget. At the highest-scoring locations, 2D token features replace the spatial features recovered from the 1D representation, and a shared decoder reconstructs the image from the resulting mixed features. 
On ImageNet $256\times256$, MixTok achieves highly competitive
reconstruction and generation performance, with an rFID of 1.195
and a gFID of 1.95 at respective average budgets of approximately
128 and 131 content tokens per image.

%% file: sections/introduction_yefei.tex
Visual tokenizers compress high-dimensional images into compact latent representations that support reconstruction and provide a modeling space for downstream generators \citep{esser2021taming,yu2024titok,rombach2022ldm,peebles2023dit}. A central challenge is to allocate representational capacity efficiently: visual complexity varies both within and across images, yet compact representations must retain the information needed for faithful reconstruction and generation. This calls for tokenizers that adapt token allocation to image content.

Existing visual tokenizers typically represent an image as either a spatial grid of 2D tokens or a compact 1D latent sequence. Conventional 2D tokenizers preserve an explicit correspondence between tokens and spatial locations, enabling localized representations of structures and textures \citep{esser2021taming}. However, their uniform spatial allocation assigns the same token density to visually simple regions, such as smooth sky, as to regions containing intricate objects or textures, potentially wasting representational capacity. In contrast, 1D tokenizers aggregate information across the image into compact sequences without a one-to-one spatial correspondence \citep{yu2024titok}. Although this global aggregation improves compactness, preserving fine-grained local details within a short sequence remains challenging \citep{wang2026vltok}. These complementary limitations motivate combining a compact global representation with selectively allocated local capacity.

Beyond spatial allocation, most existing tokenizers assign the same number of tokens to all images at a given resolution, regardless of content complexity. This fixed per-image budget can allocate unnecessary capacity to simple images while limiting detail preservation in complex ones \citep{duggal2025alit}. An effective content-adaptive allocation strategy should therefore determine both where local tokens offer the greatest benefit and how many tokens to assign to each image, coordinating spatial selection and per-image allocation under a target average token budget.

To this end, we introduce MixTok, which adaptively allocates 2D tokens for local refinement. Built upon off-the-shelf 1D and 2D tokenizers, MixTok first maps a small set of 1D tokens into a spatial feature map that provides a base representation of the image. A content-adaptive router then determines where and how much to refine: a grid scorer ranks spatial locations by refinement priority, while a top-p budget controller determines the number of selected locations from the resulting score distribution. The selected 2D features refine the 1D-derived base grid, and a shared decoder reconstructs the image from the combined representation. This design retains the compactness of the global representation while allocating additional capacity to local details that require refinement.
For downstream image generation, we employ a two-stage MaskGIT~\citep{chang2022maskgit} pipeline. The first stage generates the 1D base tokens, which guide the selection of refinement locations and the number of 2D tokens to generate. The second stage generates these selected 2D tokens, after which the shared decoder produces the image from the combined representation. On ImageNet at $256\times256$, MixTok achieves an rFID of 1.195 and a gFID of 1.95 with respective average budgets of approximately 128 and 131 tokens per image, demonstrating high-quality reconstruction and generation with compact representations.

Our main contributions are as follows:
\begin{itemize}
\item We introduce MixTok, a hybrid visual tokenizer that combines compact 1D global representations with sparse 2D local refinement through spatial feature replacement and a shared decoder.
\item We develop a content-adaptive router that jointly selects refinement locations and determines per-image token counts through learned spatial scoring and top-p selection under a target average token budget.
\item We extend MixTok to downstream image generation through a two-stage MaskGIT pipeline and demonstrate its reconstruction and generation performance on ImageNet at $256\times256$.
\end{itemize}

%% file: sections/relatedwork_yefei.tex
\textbf{Visual tokenization.} Visual tokenizers map images into compact latent representations for reconstruction and downstream generation \citep{ramesh2021dalle,chang2023muse,gu2022vqdiffusion,tian2024var,hansenestruch2025vitok}. VQ-VAE~\citep{oord2017neural} learns discrete representations by quantizing encoder features into codebook entries arranged on a spatial grid. Subsequent work improves this framework through reconstruction objectives, model architectures, and quantization schemes~\citep{razavi2019vqvae2,lee2022rqvae,zhu2025simvq,zhao2025bsq}. VQGAN~\citep{esser2021taming} introduces perceptual~\citep{zhang2018lpips} and adversarial objectives~\citep{goodfellow2014gan} to improve reconstruction quality, while ViT-VQGAN~\citep{yu2022vitvqgan} adopts Transformer-based encoders and decoders~\citep{vaswani2017attention} with improved codebook learning. MAGVIT-v2~\citep{yu2024magvitv2} introduces lookup-free quantization based on binary latent codes, enabling larger token vocabularies. These methods improve the quality of spatial representations, but their fixed latent grids impose uniform token density across image regions.

Another line of work replaces spatial grids with compact 1D latent sequences. TiTok~\citep{yu2024titok} uses learnable latent tokens to aggregate image information into a short discrete sequence. TA-TiTok~\citep{kim2025tatitok} extends this approach with text-conditioned decoding and a simplified single-stage training procedure. In the continuous setting, MAETok~\citep{chen2025maetok} uses masked modeling and auxiliary feature prediction to learn compact, semantically structured representations for diffusion models. Beyond sequence compactness, SelfBootTok~\citep{chi2026selfboottok} separates global and local token groups and learns to predict local details from global tokens, allowing downstream generators to model only the global tokens. However, recovering local content solely from a compact global representation may remain challenging when image-specific details are not adequately encoded in that representation. MixTok addresses this limitation by retaining additional 2D tokens at selected spatial locations to complement the compact 1D representation.     

\textbf{Adaptive visual tokenization.} Adaptive visual tokenization seeks to balance representation size and reconstruction quality by varying token counts or spatial allocation. One direction supports flexible per-image token budgets. FlexTok~\citep{bachmann2025flextok} learns an ordered, coarse-to-fine 1D sequence that supports reconstruction at different chosen lengths. Such flexibility enables variable token budgets but does not, by itself, specify how to select a budget for each image. ALIT~\citep{duggal2025alit} adapts representation length through recurrent token allocation, progressively adding latent tokens to improve reconstruction. CAT~\citep{shen2025cat} uses caption-based complexity assessment to select an image-dependent compression ratio.
A related direction adapts the spatial distribution or support of tokens. DQ-VAE~\citep{huang2023dqvae} varies coding granularity according to regional information density, assigning fewer codes to smooth regions and more to regions containing fine details. GPSToken~\citep{zhang2025gpstoken} represents image regions using parameterized 2D Gaussians with associated texture features, allowing token positions and spatial supports to adapt to image content. However, spatial adaptation alone does not adjust the image-level token budget to the overall information content of each image. In contrast, MixTok jointly determines where to allocate 2D refinement tokens and how many each image requires, assigning additional capacity according to the refinement needs of the 1D representation while maintaining a target average token budget.

%% file: sections/preliminaries_yefei.tex
\label{sec:preliminary-vq}

Visual tokenizers map images into compact latent representations for
reconstruction and downstream generation. Given an image
$x\in\mathbb{R}^{H\times W\times3}$, an encoder $E$ transforms it into
continuous latent features $Z=E(x)$. A vector quantizer then maps these
features into discrete codebook indices and their corresponding quantized
features:
\begin{equation}
    (T,F)=\operatorname{VQ}_{\mathcal{C}}(Z),
\end{equation}
where $T$ denotes the discrete token representation modeled by downstream
generative models, and $F$ denotes the quantized features used for image
reconstruction.

\paragraph{2D grid tokenization.}
A 2D tokenizer preserves the spatial correspondence between tokens and image
locations. Its encoder produces a latent grid
$Z_{\mathrm{2D}}=E_{\mathrm{2D}}(x)$, which is quantized into discrete tokens
and spatially organized features:
\begin{equation}
    (T_{\mathrm{2D}},F^{q}_{\mathrm{2D}})
    =
    \operatorname{VQ}_{\mathcal{C}_{\mathrm{2D}}}
    (Z_{\mathrm{2D}}).
\end{equation}
Here, $T_{\mathrm{2D}}$ denotes the spatial token grid and
$F^{q}_{\mathrm{2D}}$ denotes the corresponding quantized feature grid.
The pretrained post-quantization projection maps it to the decoder input
space:
\begin{equation}
    F_{\mathrm{2D}}
    =
    P_{\mathrm{2D}}(F^{q}_{\mathrm{2D}})
    \in
    \mathbb{R}^{h\times w\times d_{\mathrm{2D}}}.
\end{equation}
The resulting feature grid preserves spatial organization and can be used
for local refinement and decoding.

\paragraph{1D sequence tokenization.}
A 1D tokenizer compresses an image into a compact latent sequence without
explicit correspondence between tokens and spatial locations. Following
TiTok~\citep{yu2024titok}, the encoder produces a sequence of latent features:
\begin{equation}
    Z_{\mathrm{1D}}
    =
    E_{\mathrm{1D}}(x)
    \in
    \mathbb{R}^{L\times d_{\mathrm{1D}}},
\end{equation}
which is quantized into discrete tokens and their corresponding features:
\begin{equation}
    (T_{\mathrm{1D}},F_{\mathrm{1D}})
    =
    \operatorname{VQ}_{\mathcal{C}_{\mathrm{1D}}}
    (Z_{\mathrm{1D}}).
\end{equation}
Here, $T_{\mathrm{1D}}$ denotes the compact token sequence and
$F_{\mathrm{1D}}$ denotes its corresponding quantized features.

%% file: sections/42_43_lichenge.tex
\begin{figure}[t]
    \centering
    \includegraphics[width=\linewidth]{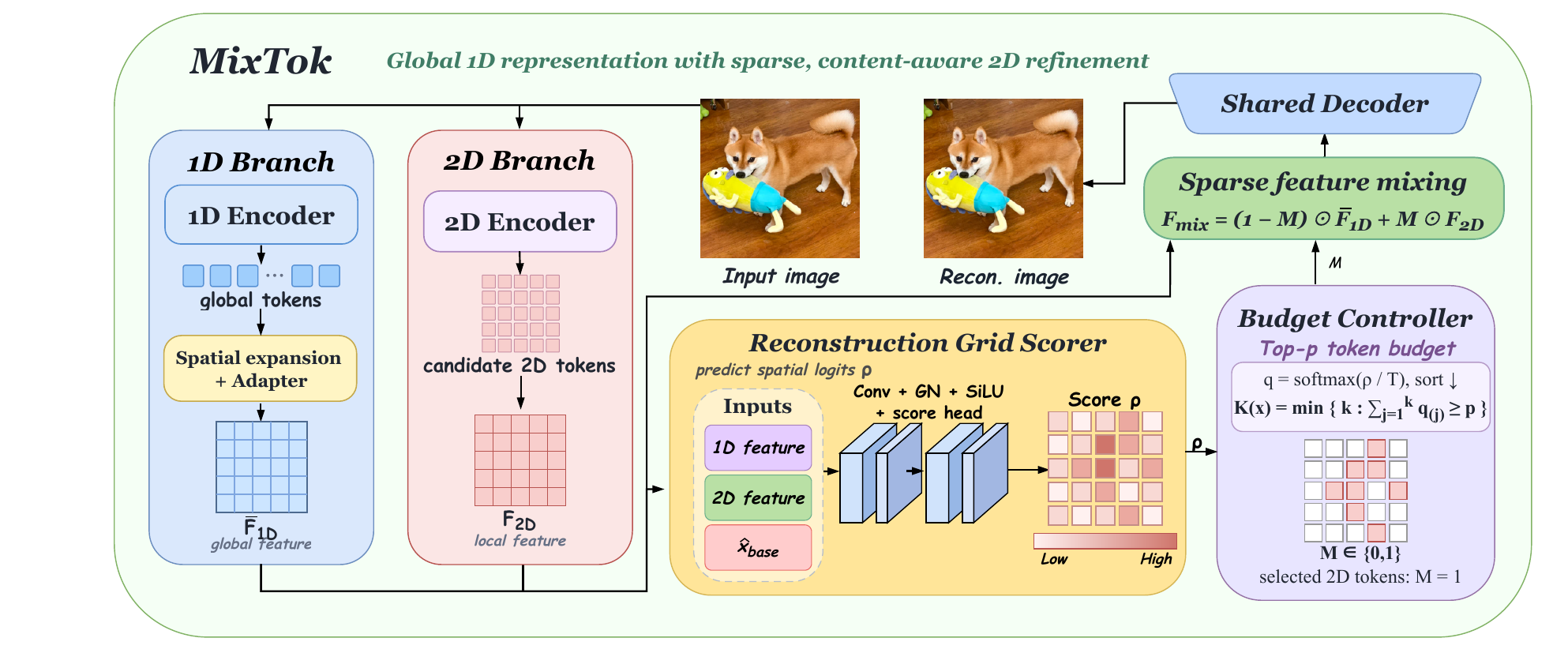}
\caption{\textbf{Overview of MixTok.}
Compact 1D tokens provide a base feature grid, while
content-adaptive top-$p$ selection allocates 2D refinement
tokens across spatial locations.
The selected 2D features replace the corresponding base
features, and a shared decoder reconstructs the image
from the mixed grid.}
    \label{fig:MixTok-overview}
\end{figure}

\subsection{Hybrid Visual Representation}
\label{sec:hybrid-representation}

To combine the compactness of 1D representations with the spatial detail of 2D grids, MixTok supplements a short 1D token sequence with selected 2D tokens for local refinement.

Formally, given an image $x$, the 1D and 2D branches produce a length-$L$
token sequence $T_{\mathrm{1D}}$ and an $h\times w$ token grid
$T_{\mathrm{2D}}$, respectively. The 2D grid provides
$N=hw$ candidate refinement locations. Let
$\mathcal{R}=\{r_1,\ldots,r_K\}$ denote the selected locations, where each
$r_k\in\{1,\ldots,N\}$ indexes a spatial position under a fixed flattening
order. MixTok retains the complete 1D sequence together with the selected 2D
tokens:
\begin{equation}
    \mathcal{T}_{\mathrm{MixTok}}(x)
    =
    \left(
        T_{\mathrm{1D}},
        \{
        (r_k,t_{r_k}^{\mathrm{2D}})
        \}_{k=1}^{K}
    \right),
    \qquad B=L+K ,
\end{equation}
where $t_{r_k}^{\mathrm{2D}}$ denotes the 2D token at location $r_k$ and
$B$ denotes the number of retained content tokens. The selected locations are
stored as side information to recover their spatial positions during decoding
and are not counted as content tokens.

To combine the two token types, MixTok maps their quantized features into a
shared decoder feature space. Since 1D tokens do not preserve explicit spatial
correspondence, their quantized features $F_{\mathrm{1D}}$ are first expanded
into a spatial feature grid using the spatial query grid $S$:
\begin{equation}
    \widetilde{F}_{\mathrm{1D}}
    =
    G_{\mathrm{1D}}(F_{\mathrm{1D}},S).
\end{equation}
A learnable feature adapter then projects the expanded features into the
decoder input space:
\begin{equation}
    \bar{F}_{\mathrm{1D}}
    =
    P_{\mathrm{1D}}(\widetilde{F}_{\mathrm{1D}})
    \in
    \mathbb{R}^{h\times w\times d_{\mathrm{2D}}}.
\end{equation}

In contrast, the decoder-aligned features from the 2D branch,
  $F_{\mathrm{2D}}\in\mathbb{R}^{h\times w\times d_{\mathrm{2D}}}$,
  preserve spatial organization and match the decoder input dimension.
  Therefore, $\bar{F}_{\mathrm{1D}}$ and $F_{\mathrm{2D}}$ can be combined
  within the same feature space.

MixTok constructs a complete decoder input grid by replacing the 1D-derived
feature with the corresponding 2D feature at each selected refinement
location:
\begin{equation}
    F_{\mathrm{mix},i}
    =
    \begin{cases}
        F_{\mathrm{2D},i}, & i\in\mathcal{R},\\
        \bar{F}_{\mathrm{1D},i}, & i\notin\mathcal{R},
    \end{cases}
    \qquad
    \hat{x}
    =
    D(F_{\mathrm{mix}}),
    \label{eq:MixTok-mixed-decoding}
\end{equation}
where $i\in\{1,\ldots,N\}$ indexes the spatial locations of the feature grid
under the same flattening order used above. 

\subsection{Content-Adaptive Token Allocation}
\label{sec:content-adaptive-allocation}

The benefit of local refinement varies across spatial regions and images.
When the 1D representation already captures a region well,
allocating additional 2D tokens provides limited improvement. 
For reconstruction, MixTok therefore uses a content-adaptive router comprising
a Reconstruction Grid Scorer and a Budget Controller.
The former scores candidate locations, and the latter uses these scores
to determine the refinement locations and token budget for each image.

\paragraph{Reconstruction grid scoring.}
The Reconstruction Grid Scorer predicts refinement priorities from
the aligned features of both branches and the base reconstruction
$\hat{x}_{\mathrm{base}}=D(\bar{F}_{\mathrm{1D}})$:
\begin{equation}
    \rho^{\mathrm{rec}}(x)
    =
    \Phi_{\mathrm{grid}}^{\mathrm{rec}}
    \left(\bar{F}_{\mathrm{1D}},F_{\mathrm{2D}},
    \hat{x}_{\mathrm{base}}\right),
\end{equation}
where $\rho_i^{\mathrm{rec}}(x)$ denotes the refinement score
at location $i$.
Sorting these scores in descending order gives
$\pi_1(x),\ldots,\pi_N(x)$, where $\pi_k(x)$ denotes the
$k$-th highest-scoring location.
The scorer receives auxiliary supervision from local reconstruction-error reductions between the 1D base and full 2D reconstructions, providing a proxy for refinement benefit. It is also optimized through the mixed reconstruction loss using a differentiable selection surrogate as shown in Appendix \ref{app:scorer-supervision}.

\paragraph{Budget control.}
The Budget Controller uses top-$p$ selection~\citep{holtzman2020nucleus}
to determine how many 2D tokens to retain from the score distribution
predicted by the Reconstruction Grid Scorer.
We first normalize the scores using a softmax with temperature $T>0$:
\begin{equation}
    q_i(x)
    =
    \frac{\exp(\rho_i^{\mathrm{rec}}(x)/T)}
    {\sum_{j=1}^{N}\exp(\rho_j^{\mathrm{rec}}(x)/T)}.
    \label{eq:MixTok-score-normalization}
\end{equation}
Given a threshold $p\in(0,1]$, we select the shortest prefix of the
ranking whose cumulative probability reaches $p$:
\begin{equation}
    K(x)
    =
    \min
    \left\{
        k\in\{1,\ldots,N\}
        \;\middle|\;
        \sum_{j=1}^{k}q_{\pi_j(x)}(x)\ge p
    \right\}.
    \label{eq:MixTok-top-p-budget}
\end{equation}
The refinement set and total number of retained content tokens are
\begin{equation}
    \mathcal{R}(x)
    =
    \{\pi_1(x),\ldots,\pi_{K(x)}(x)\},
    \qquad
    B(x)=L+K(x).
    \label{eq:MixTok-adaptive-selection}
\end{equation}
This rule allocates fewer refinement tokens when the probability mass
is concentrated on a small number of locations and more tokens when
it is more dispersed. The threshold $p$ controls the trade-off between
refinement coverage and token budget. We calibrate $p$ on a subset of training images to match the target average token budget and keep it fixed during evaluation, while allowing token counts to vary across images.

\paragraph{Training objective.}
Following common practice~\citep{sun2024llamagen,yu2022vitvqgan}, MixTok is
optimized with a weighted combination of reconstruction, perceptual, and
adversarial losses:
\begin{equation}
    \mathcal{L}_{\mathrm{MixTok}}
    =
    \lambda_{\mathrm{rec}}
    \|x-\hat{x}\|_1
    +
    \lambda_{\mathrm{per}}
    \mathcal{L}_{\mathrm{LPIPS}}(x,\hat{x})
    +
    \lambda_{\mathrm{adv}}
    \mathcal{L}_{\mathrm{GAN}},
    \label{eq:MixTok-loss}
\end{equation}
where $x$ and $\hat{x}$ denote the input and reconstructed images,
respectively, and
$\lambda_{\mathrm{rec}}$, $\lambda_{\mathrm{per}}$, and
$\lambda_{\mathrm{adv}}$ control the weights of different loss terms.
Detailed loss definitions and weights are provided in
  Appendix~\ref{app:reconstruction-losses}.

\subsection{Image Generation with MixTok}
\label{sec:generation-with-MixTok}

\begin{figure}[H]
    \centering
    \includegraphics[width=\linewidth]{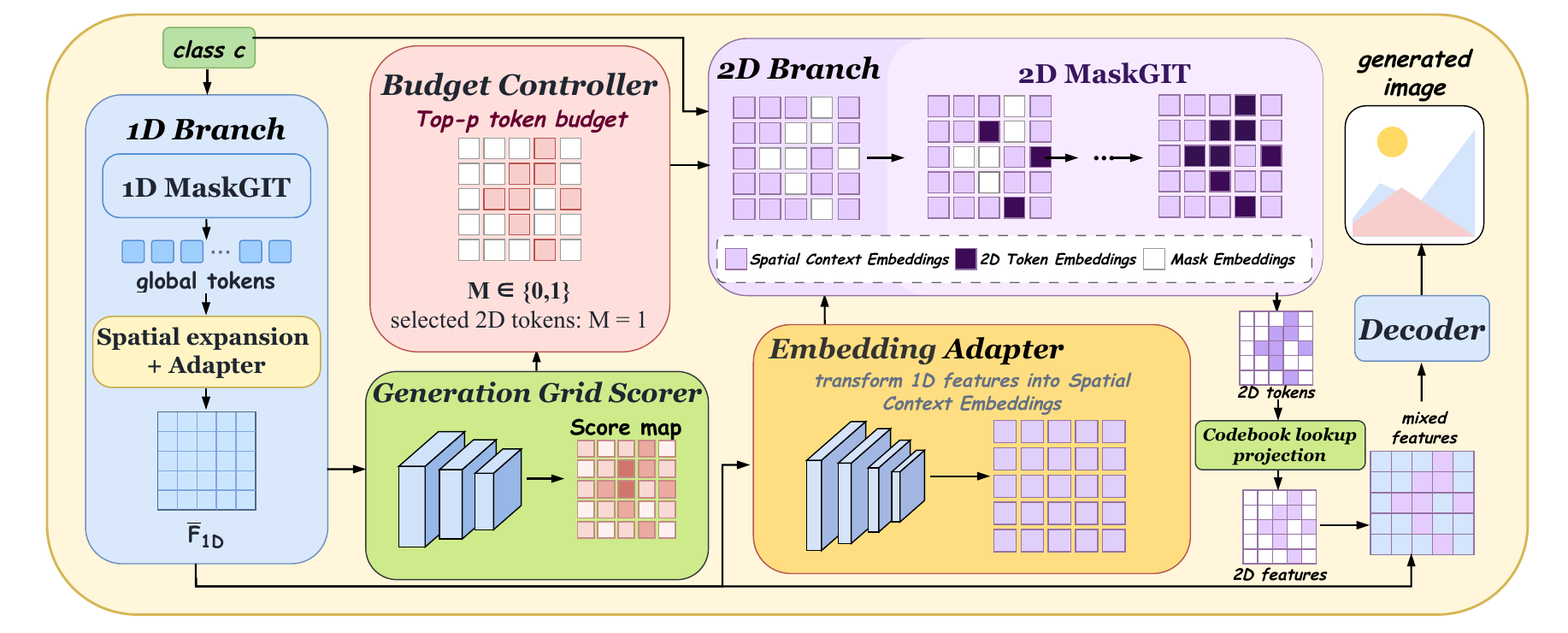}
\caption{\textbf{Two-stage masked generation with MixTok.}
A 1D MaskGIT generates global tokens that guide refinement
location selection and provide spatial context for a 2D MaskGIT.
Features obtained from the generated 2D tokens replace the
corresponding base features before shared decoding.}
    \label{fig:MixTok-maskgit}
\end{figure}

MixTok generates its hybrid representation in two stages:
global 1D tokens establish a base representation, which
then guides the selection and generation of sparse 2D
refinement tokens (Figure~\ref{fig:MixTok-maskgit}).
Given a class condition $c$, a pretrained TiTok MaskGIT~\citep{yu2024titok}
generates a length-$L$ sequence $\hat{T}_{\mathrm{1D}}$.
Codebook lookup, spatial expansion, and feature alignment
convert this sequence into the base grid
$\bar{F}_{\mathrm{1D}}$, following Section \ref{sec:hybrid-representation}.

\paragraph{Refinement location selection.}
During generation, the 2D features required by the
Reconstruction Grid Scorer are unavailable.
We therefore introduce a Generation Grid Scorer that
predicts refinement priorities from the 1D base grid alone:
$\rho^{\mathrm{gen}}=
\Phi_{\mathrm{grid}}^{\mathrm{gen}}(\bar{F}_{\mathrm{1D}})
\in\mathbb{R}^{N}$.
The Generation Grid Scorer is trained by distilling the frozen
Reconstruction Grid Scorer, learning to predict refinement priorities
using only the 1D base features.
The Budget Controller applies the top-$p$ rule in
Section~\ref{sec:content-adaptive-allocation} to the outputs of the
Generation Grid Scorer, yielding the refinement set $\mathcal{R}$
and its size $K$.

\paragraph{Spatial context embeddings.}
The base grid lies in the image decoder's feature space, whereas
the 2D generator processes inputs in its token embedding space.
To use the base representation as a generation condition, we introduce
an Embedding Adapter $A$ that maps the base features to spatial context
embeddings:
\begin{equation}
    H_{\mathrm{ctx}}
    =
    A(\bar{F}_{\mathrm{1D}}).
    \label{eq:generation-context}
\end{equation}
The Generation Grid Scorer and Budget Controller determine which
locations to generate, while the Embedding Adapter supplies continuous
context at the remaining locations. The initial input embeddings are
\begin{equation}
    e_i^{(0)}
    =
    \begin{cases}
        E_{\mathrm{tok}}([\mathrm{MASK}]), & i\in\mathcal{R},\\
        H_{\mathrm{ctx},i}, & i\notin\mathcal{R},
    \end{cases}
    \label{eq:generation-masked-input}
\end{equation}
where $E_{\mathrm{tok}}$ is the 2D generator's token embedding table.
Positional embeddings are added before the Transformer.

\paragraph{Sparse 2D generation and decoding.}
A fine-tuned Halton-MaskGIT~\citep{besnier2025halton} generates 2D tokens
only in $\mathcal{R}$, conditioned on class $c$ and the context embeddings
at unselected positions.
Sampling follows the Halton order restricted to $\mathcal{R}$.
Each generated token replaces its mask embedding with
$E_{\mathrm{tok}}(t_i)$, while the context embeddings at unselected
positions remain fixed. The generated 2D tokens are converted to decoder-aligned features and
combined with the base grid using Eq.~\eqref{eq:MixTok-mixed-decoding}
to produce the final image.

\paragraph{Generation training.}
To train the Embedding Adapter $A$, we decode the base features into
images and re-encode these base reconstructions with the 2D tokenizer.
The resulting tokens are looked up in the frozen pretrained Halton-MaskGIT~\citep{besnier2025halton}
token embedding table to obtain regression targets for $A$.
We then freeze $A$ and fine-tune the Halton generator using the
2D tokens of the original training images as targets, applying
cross-entropy only to masked positions within $\mathcal{R}$.

%% file: sections/experiment.tex
\subsection{Experimental Setup}
\label{sec:experimental-setup}

\paragraph{Dataset and metrics.}
We train on ImageNet~\citep{deng2009imagenet,russakovsky2015ilsvrc}
$256\times256$ and evaluate reconstruction on all 50,000 validation images using rFID~\citep{heusel2017fid} and peak signal-to-noise ratio (PSNR). Released tokenizer checkpoints are reevaluated with common preprocessing and metric computation.
We report gFID on 50,000 class-balanced samples per sampling seed using the ADM
evaluator~\citep{dhariwal2021adm}. 

\paragraph{Model configurations.}
Our main configuration combines TiTok-L-32~\citep{yu2024titok}
(32 1D tokens) with LlamaGen VQ-16~\citep{sun2024llamagen}
(a $16\times16$ 2D grid and the shared image decoder).
For reconstruction, the fixed variant retains 96 2D tokens and
the top-$p$ variant uses approximately 96.1 on average.
We also evaluate transferability using
ALIT~\citep{duggal2025alit} and its VQGAN backbone.

\paragraph{Training strategy.}
  Starting from pretrained tokenizers, we first align the 1D features
  to the shared decoder space, then train with a fixed refinement
  budget before fine-tuning with top-$p$ allocation.
  The 1D encoder and quantizer remain frozen throughout tokenizer training.
  Detailed training and evaluation settings are provided in
  Appendix~\ref{app:appendix}.

\subsection{Image Reconstruction}
\label{sec:image-reconstruction}

\begin{figure}[H]
    \centering
    \includegraphics[width=\textwidth]{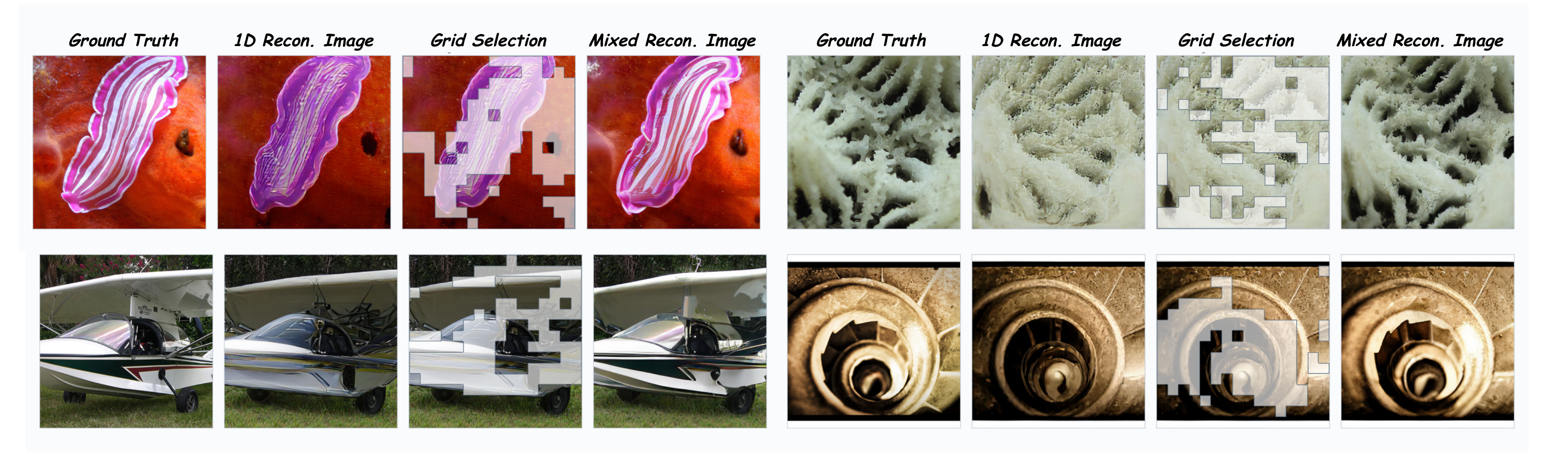}
\caption{\textbf{Qualitative reconstruction results.}
Selective 2D refinement recovers fine details and sharpens
structures that are blurred or missing in the 1D base
reconstructions, while retaining a compact token representation.}
    \label{fig:reconstruction-qualitative}
\end{figure}

\paragraph{Reconstruction comparison.}
  As shown in Table~\ref{tab:reconstruction-baselines}, MixTok achieves
  an rFID of 1.1950 and a PSNR of 19.51\,dB with approximately 128
  content tokens on average. At a comparable token budget, it improves
  rFID over both TiTok variants and One-D-Piece, while remaining
  competitive with FlexTok and DetailFlow.
  Figure~\ref{fig:reconstruction-qualitative} shows that 2D refinement
  improves stripe patterns and curved boundaries over the 1D base
  reconstructions.

\begin{table}[H]
    \centering
    \caption{Reconstruction quality and average content-token count. Token counts are rounded to the nearest integer. Baseline configurations are detailed in Appendix~\ref{app:reconstruction-baselines}. Estimated bitrates are reported in Appendix~\ref{app:bitrate-accounting}.}
    \label{tab:reconstruction-baselines}
    \setlength{\tabcolsep}{6pt}
    \begin{tabular}{lccc}
        \toprule
        Tokenizer & Avg. tokens & rFID $\downarrow$ & PSNR (dB) $\uparrow$ \\
        \midrule
        TiTok S-128~\citep{yu2024titok} & 128 & 1.7222 & 17.72 \\
        TiTok BL-128 VQ~\citep{kim2025tatitok} & 128 & 1.5063 & 19.83 \\
        One-D-Piece L-256~\citep{miwa2025onedpiece} & 128 & 1.4406 & 17.71 \\
        FlexTok d18-d28~\citep{bachmann2025flextok} & 128 & 1.1422 & 16.76 \\
        DetailFlow-16~\citep{liu2025detailflow} & 128 & 1.1370 & 19.46 \\
        \midrule
        LlamaGen VQ-16~\citep{sun2024llamagen} & 256 & 2.1872 & 20.78 \\
        MaskGIT-VQGAN~\citep{chang2022maskgit} & 256 & 2.1304 & 18.15 \\
        \midrule
        \textbf{MixTok (ours)} & 128 & 1.1950 & 19.51 \\
        \bottomrule
    \end{tabular}
\end{table}

\paragraph{Transferability across tokenizer backbones.}
For the ALIT--VQGAN combination, we use 64 1D tokens and initialize
MixTok-ALIT from the released ALIT-128 checkpoint. We train with a fixed
budget of 64 2D tokens for 20 epochs, followed by 5 epochs of top-$p$
training. At evaluation, we retain exactly 64 2D tokens, giving 128
content tokens in total to match the ALIT-128 baseline.

\begin{table}[H]
    \centering
    \caption{Transferability to ALIT--VQGAN at a fixed budget of 128 content tokens.}
    \label{tab:alit-transferability}
    \setlength{\tabcolsep}{7pt}
    \begin{tabular}{lccc}
        \toprule
        Tokenizer & Tokens & rFID $\downarrow$ & PSNR (dB) $\uparrow$ \\
        \midrule
        ALIT-128 & 128 & 2.9453 & 18.71 \\
        MixTok-ALIT & 128 & \textbf{2.4720} & \textbf{18.76} \\
        \bottomrule
    \end{tabular}
\end{table}

Table~\ref{tab:alit-transferability} shows that, at the same
  128-token budget, MixTok-ALIT uses 64 1D and 64 2D tokens and
  improves rFID from 2.9453 to 2.4720 and PSNR from 18.71 to
  18.76\,dB over ALIT-128. These results support the hybrid
  representation at a matched content-token budget and demonstrate
  transferability beyond TiTok--LlamaGen.

\subsection{Image Generation}
\label{sec:image-generation}

\begin{figure}[H]
    \centering
    \includegraphics[width=0.82\linewidth]{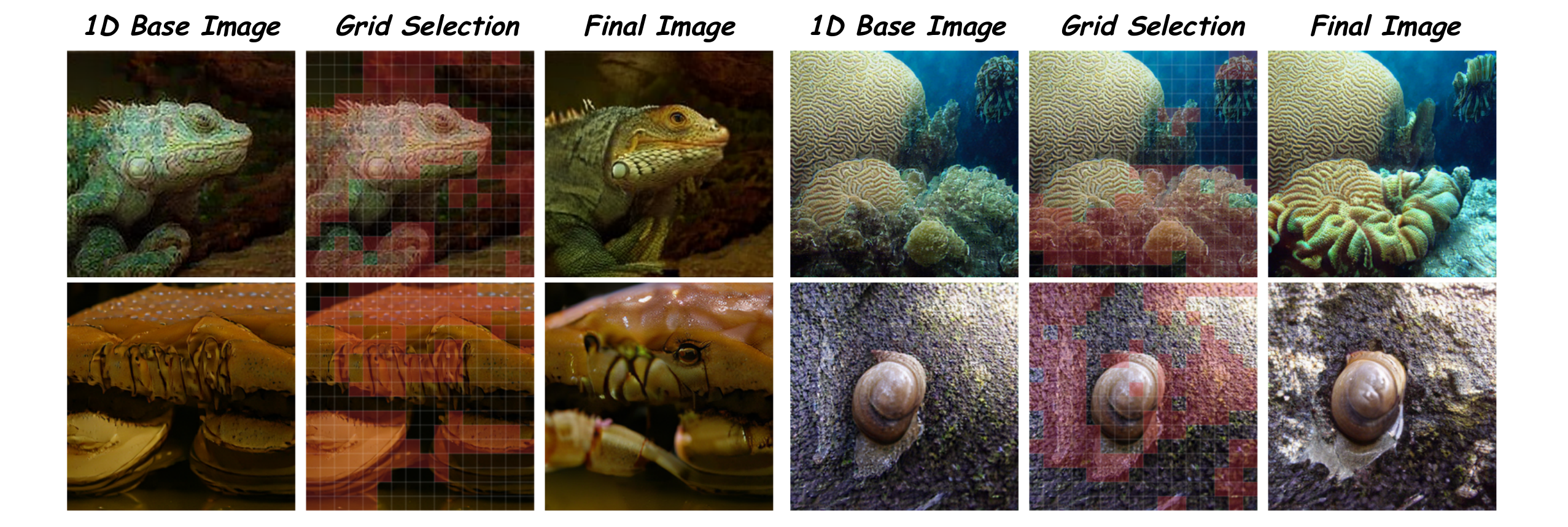}
    \caption{Qualitative class-conditional generation with MixTok.}
    \label{fig:generation-qualitative}
\end{figure}

As shown in Table~\ref{tab:generation-baselines}, MixTok achieves a gFID
of 1.95 with approximately 131 tokens. At a similar token budget,
it improves gFID over TiTok-S-128 from 2.00 to 1.95 while reducing
sampling rounds from 64 to 16. Compared with DetailFlow-32 at a similar
parameter count, MixTok achieves a lower gFID (1.95 vs.\ 2.63) with
approximately half as many tokens and fewer rounds (16 vs.\ 39).
Under our single-image latency protocol, MixTok is $3.77\times$ and
$2.48\times$ faster than these TiTok and DetailFlow configurations,
respectively. Starting from the same generated 1D tokens, sparse 2D refinement improves gFID from 4.33 to 1.95 with eight additional sampling rounds, demonstrating its contribution to generation quality.
Figure~\ref{fig:generation-qualitative} illustrates improved local
  details after 2D refinement during generation.
  Quantitative results and sampling-round ablations are provided in
  Table~\ref{tab:generation-seeds} and Appendix~\ref{app:stage2-steps}.

\begin{table}[H]
    \centering
    \caption{Class-conditional ImageNet generation. Unmarked gFID values
    are from our ADM reevaluations, while $^*$ denotes an author-reported result.
    Latency and FLOPs are measured on
    one RTX 5090 at batch size 1. Configuration and measurement details are provided in Appendix~\ref{app:generation-efficiency}.}
    \label{tab:generation-baselines}
    \small
    \setlength{\tabcolsep}{3pt}
    \renewcommand{\arraystretch}{1.15}
    \begin{tabularx}{\linewidth}{@{}>{\raggedright\arraybackslash}Xrrrrrr@{}}
        \toprule
        Generator & Params & Tokens & Steps & gFID $\downarrow$
        & \shortstack{Latency $\downarrow$\\(ms)}
        & \shortstack{FLOPs $\downarrow$\\(T/image)} \\
        \midrule
        MaskGIT~\citep{chang2022maskgit} & 227M & 256 & 8 & 6.06$^*$ & --- & --- \\
                AdaGen-MaskGIT-L~\citep{ni2026adagen} & 206M & 256 & 16 & 2.4883 & 372.0 & 3.623 \\
        TiTok-S-128~\citep{yu2024titok} & 295M & 128 & 64 & 1.9994 & 1018.4 & 9.949 \\
        Halton-MaskGIT-L~\citep{besnier2025halton} & 480M & 576 & 32 & 2.5143 & 591.5 & 26.690 \\
        One-D-Piece-L-256~\citep{miwa2025onedpiece} & 177M & 256 & 16 & 2.4144 & 350.1 & 3.526 \\
        LPD-L-256~\citep{zhang2026lpd} & 337M & 256 & 20 & 2.3863 & \textbf{215.5} & 0.879 \\
        LlamaGen-L~\citep{sun2024llamagen} & 343M & 256 & 256 & 3.6803 & 4951.8 & \textbf{0.599} \\
        DetailFlow-32~\citep{liu2025detailflow} & 326M & 256 & 39 & 2.6315 & 669.9 & 5.627 \\
        \midrule
        \textbf{MixTok (ours)} & 321M & 130.66 & $8+8$
        & $\mathbf{1.9511}$ & 270.1 & 1.452 \\
        \bottomrule
    \end{tabularx}
\end{table}

\subsection{Ablation Studies}
\label{sec:ablation-studies}

\paragraph{Grid scorer.}
To evaluate whether learned scores identify effective refinement locations, we compare the Reconstruction and Generation Grid Scorers with random, uniform, and gradient-entropy rankings. We hold each image's top-$p$ budget $K(x)$ from the Reconstruction Grid Scorer fixed across all strategies, isolating spatial ranking from budget allocation. Uniform ranking follows a predefined spatially dispersed order, while gradient-entropy ranking uses the Shannon entropy of grayscale Sobel gradient magnitudes. As shown in Table~\ref{tab:grid-selector-ablation}, the Reconstruction Grid Scorer achieves the best rFID of 1.1950. Using only 1D base features, the Generation Grid Scorer also outperforms all tested heuristics, achieving an rFID of 1.3479 and a PSNR of 18.4776. These results support learned spatial selection and show that distillation preserves useful refinement priorities without access to 2D features.

\begin{table}[H]
    \centering
        \caption{Grid Scorer ablation with identical per-image budgets. Random ranking reports mean $\pm$ standard deviation over three seeds; the other rankings are deterministic.}
    \label{tab:grid-selector-ablation}
    \setlength{\tabcolsep}{8pt}
    \begin{tabular}{lcc}
        \toprule
        Selection strategy & rFID $\downarrow$ & PSNR $\uparrow$ \\
        \midrule
        Reconstruction Grid Scorer & \textbf{1.1950} & \textbf{19.5074} \\
        Generation Grid Scorer & 1.3479 & 18.4776 \\
        Random & $1.5460 \pm 0.0093$ & $17.2147 \pm 0.0010$ \\
        Uniform & 1.6604 & 17.1749 \\
        Gradient entropy & 1.4012 & 18.0077 \\
        \bottomrule
    \end{tabular}
\end{table}

\paragraph{Budget controller.}
To assess whether matching refinement budgets to image content improves reconstruction, we compare top-$p$ allocation with fixed $K=96$ and shuffled budgets, using the same Reconstruction Grid Scorer. The shuffled control randomly permutes the top-$p$ budgets across images, preserving their distribution and mean while removing their correspondence with image content. Each strategy refines the top-ranked locations for its assigned budget. As shown in Table~\ref{tab:k-selector-ablation}, top-$p$ allocation achieves the best rFID and PSNR at a comparable average budget. Shuffling the budgets degrades both metrics, supporting the benefit of content-dependent allocation beyond merely varying token counts.

\begin{table}[H]
    \centering
    \caption{Budget Controller ablation with a fixed Grid Scorer. The shuffled control preserves the top-p budget histogram and reports mean $\pm$ standard deviation over three permutations.}
    \label{tab:k-selector-ablation}
    \setlength{\tabcolsep}{7pt}
    \begin{tabular}{lccc}
        \toprule
        Allocation strategy & Avg. $K$ & rFID $\downarrow$ & PSNR $\uparrow$ \\
        \midrule
        Fixed $K=96$ & 96.0000 & 1.2052 & 19.4591 \\
        Top-p dynamic $K$ & 96.1093 & \textbf{1.1950} & \textbf{19.5074} \\
        Shuffled $K$ & 96.1093 & $1.2236 \pm 0.0068$ & $19.4181 \pm 0.0005$ \\
        \bottomrule
    \end{tabular}
\end{table}

%% file: sections/conclusion_and_future_work.tex
We presented MixTok, a hybrid visual tokenizer that combines compact 1D global tokens with sparse 2D tokens to preserve local details. Its content-adaptive router determines both where to refine and how many tokens to allocate to each image under a target average budget. The resulting representation supports reconstruction and two-stage masked generation, achieving an rFID of 1.195 and a gFID of 1.95 on ImageNet at average budgets of approximately 128 and 131 content tokens, respectively. Ablations and experiments across tokenizer backbones support the effectiveness of this design. These results demonstrate the value of combining global compression with content-adaptive local refinement for compact, high-quality visual representations.

\paragraph{Limitations and future work.}
Although MixTok retains and generates fewer discrete content tokens, reconstruction still requires dense 2D encoding, and the second-stage generator operates on the full spatial grid. Evaluation is currently limited to ImageNet at $256\times256$ and a narrow range of average token budgets. Future work will explore sparse computation and extend adaptive global--local token allocation to broader budgets, higher resolutions, and more diverse visual data.

%% file: sections/ai_use.tex
We used generative AI tools, primarily Codex, to assist with method
implementation and minor refinements, experimental design and parameter
selection, debugging, data processing, experiment execution, and metric
aggregation. We also used these tools for translation, manuscript revision,
literature search, and reference checking. The main methodological advances
were proposed by the authors. The authors reviewed and revised the
AI-assisted code, checked experimental logs and metrics, consulted original
references, and reviewed the manuscript. The authors take full responsibility
for the final content of this work, including its claims, code, and reported
results.

%% file: sections/ethics_statement.tex
This study uses ImageNet and publicly available pretrained models and does
not involve new data collection or studies with human participants. Like
other image generation systems, the method may inherit biases from training
data and could be misused to produce misleading imagery. These risks should
be considered when deploying downstream systems.

%% file: sections/reproducibility_statement.tex
Our method and experimental details are provided in
Sections~\ref{sec:method} and~\ref{sec:experimental-setup}
and Appendix~\ref{app:appendix}.
Source code and configuration files for reconstruction and generation are
available at \url{https://github.com/LCGLiChenge/MixTok}.
The repository documents the required datasets, pretrained checkpoints,
and other external assets.
The project page is available at
\url{https://lcglichenge.github.io/MixTok-page/}.

%% file: sections/appendix.tex
\subsection{Implementation Details}
\label{app:implementation}

The settings below describe the main TiTok--LlamaGen experimental
configuration. Spatial expansion $G_{\mathrm{1D}}$, feature alignment
$P_{\mathrm{1D}}$, and the generation Embedding Adapter $A$ are separate
modules with different output spaces.

\paragraph{Tokenizer modules.}
We use TiTok-L-32~\citep{yu2024titok} with a 4,096-entry codebook of
12-dimensional vectors and LlamaGen VQ-16~\citep{sun2024llamagen} with a
16,384-entry codebook of 8-dimensional vectors. Spatial expansion
$G_{\mathrm{1D}}$ is initialized from the TiTok decoder backbone and
produces a $16\times16$ grid. The feature adapter $P_{\mathrm{1D}}$
uses a $1\times1$ projection to 256 channels, two residual convolutional
blocks, and a final $1\times1$ projection. Each residual block contains
two $3\times3$ convolutions with GroupNorm and SiLU. The 2D branch's
post-quantization projection $P_{\mathrm{2D}}$ maps its 8-dimensional
codebook vectors to the same 256-channel decoder space.

\paragraph{Grid Scorers.}
The Reconstruction Grid Scorer projects $\bar F_{\mathrm{1D}}$,
$F_{\mathrm{2D}}$, their difference, and its absolute value through
separate $1\times1$ convolutions. The base reconstruction is average-pooled
to the grid resolution and processed by two $3\times3$ convolutions
with an intermediate SiLU. These five outputs are summed with a learned
spatial embedding. Three blocks of $3\times3$ convolution, eight-group
GroupNorm, and SiLU precede a $1\times1$ score head. All hidden features
have 128 channels. The Generation Grid Scorer uses the same hidden width,
three-block trunk, and score head, with a single projection of
$\bar F_{\mathrm{1D}}$ and a learned spatial embedding as its input.

\paragraph{Generation modules.}
The Embedding Adapter $A$ adds a convolutional residual to the
256-channel base grid and projects it to 768 channels. The residual
branch uses two $3\times3$ convolutions with 128 hidden channels and SiLU,
followed by a zero-initialized $1\times1$ convolution to 256 channels.
A final $1\times1$ projection produces the spatial context embeddings.
The 2D generator is initialized from the pretrained Halton-MaskGIT
base model~\citep{besnier2025halton}, with 12 Transformer layers,
12 attention heads, hidden dimension 768, feed-forward dimension 3,072,
and dropout 0.1. It retains the full spatial sequence and one register
token; sparse refinement restricts the prediction targets and token
updates to $\mathcal R$.

\subsection{Reconstruction Training Details}
\label{app:reconstruction-training}

\subsubsection{Training Schedule}
\label{app:reconstruction-schedule}

Table~\ref{tab:appendix-reconstruction-training} summarizes the released
three-stage recipe. The 1D encoder and quantizer remain frozen throughout.
Alignment pretraining optimizes only $G_{\mathrm{1D}}$ and
$P_{\mathrm{1D}}$, with the complete 2D tokenizer frozen. Fixed-budget
training begins with 0.5 epoch of Reconstruction Grid Scorer optimization,
then jointly optimizes spatial expansion, feature alignment, the 2D
encoding path, the post-quantization projection, the image decoder,
and the scorer. Adversarial training starts at epoch 1.5, with a
0.01-epoch discriminator warm-up. Top-$p$ fine-tuning initializes from
the fixed-budget EMA model, retains the discriminator, resets the
optimizer, and enables joint and adversarial training from the start.

\begin{table}[H]
    \centering
    \caption{Reconstruction training settings for the main configuration.
    Batch sizes are global across eight GPUs. Dashes denote frozen or
    inactive modules.}
    \label{tab:appendix-reconstruction-training}
    \setlength{\tabcolsep}{3pt}
    \begin{tabular}{lccc}
        \toprule
        Setting & Alignment & Fixed budget & Top-$p$ \\
        \midrule
        Epochs & 2.5 & 20 & 6 \\
        Global batch size & 96 & 192 & 160 \\
        Refinement allocation & None & $K=96$ & $T=2,\ p=0.91$ \\
        LR: $G_{\mathrm{1D}},P_{\mathrm{1D}}$ & $5\!\times\!10^{-5}$ & $10^{-6}$ & $5\!\times\!10^{-7}$ \\
        LR: 2D encoder / pre-quantization projection & --- & $2\!\times\!10^{-7}$ & $10^{-7}$ \\
        LR: $P_{\mathrm{2D}}$, image decoder & --- & $8\!\times\!10^{-7}$ & $4\!\times\!10^{-7}$ \\
        LR: Reconstruction Grid Scorer & --- & $10^{-5}$ & $5\!\times\!10^{-6}$ \\
        LR: discriminator & --- & $5\!\times\!10^{-5}$ & $5\!\times\!10^{-5}$ \\
        LR warm-up (updates) & 1,000 & 0 & 0 \\
        Minimum LR (non-discriminator) & $5\!\times\!10^{-7}$ & 0 & 0 \\
        Weight decay: $G_{\mathrm{1D}},P_{\mathrm{1D}}$ & $10^{-4}$ & $10^{-2}$ & $10^{-2}$ \\
        EMA decay & 0.999 & 0.999 & 0.999 \\
        \bottomrule
    \end{tabular}
\end{table}

We use AdamW~\citep{loshchilov2019adamw} with
$(\beta_1,\beta_2)=(0.9,0.999)$, cosine learning-rate decay,
BF16 mixed precision, and gradient clipping at norm 1.
The discriminator uses $(0.5,0.9)$ and a constant learning rate.
Weight decay is zero for the 2D branch, scorer, and discriminator;
biases and normalization parameters are excluded from weight decay.
Tokenizer training uses bicubic resizing of the shorter image side to
256 pixels, followed by a center crop, without random cropping or flipping.
TiTok inputs lie in $[0,1]$, and LlamaGen inputs lie in $[-1,1]$.

\subsubsection{Scorer Supervision and Differentiable Selection}
\label{app:scorer-supervision}

\paragraph{Reconstruction Grid Scorer supervision.}
Let $\hat x_{\mathrm{2D}}=D(F_{\mathrm{2D}})$ be the reconstruction
using the complete 2D feature grid. We form an auxiliary supervision map
from the difference between the base and 2D reconstruction errors:
\begin{equation}
    g_i=\max\!\left\{0,\;
    \operatorname{Avg}_{u\in\Omega_i}
    \left[
    \operatorname{MSE}_{\mathrm{ch}}(\hat x_{\mathrm{base},u},x_u)
    -\operatorname{MSE}_{\mathrm{ch}}(\hat x_{\mathrm{2D},u},x_u)
    \right]\right\},
    \label{eq:appendix-gain-target}
\end{equation}
where $\Omega_i$ is the image region pooled into grid cell $i$, and
$\operatorname{MSE}_{\mathrm{ch}}$ averages over color channels.
The 37.5\% of cells with the largest $g_i$ values receive binary target 1;
the remaining cells receive 0. Binary cross-entropy with the predicted logits defines
$\mathcal L_{\mathrm{score}}$. Targets are detached from the computation
graph. This map provides local error-reduction supervision; it is a proxy
for the benefit of replacing an individual grid feature. Scorer inputs
are also detached, while the mixed reconstruction supplies gradients to
the scorer through the selection surrogate below.

\paragraph{Top-$p$ training.}
Using the probabilities and ordering from
Section~\ref{sec:content-adaptive-allocation}, let
$c_{j-1}=\sum_{\ell<j}q_{\pi_\ell}$.
The forward pass uses the hard prefix mask $m_{\pi_j}=\mathbf{1}[c_{j-1}<p]$.
For backpropagation, we use
\begin{equation}
    s_{\pi_j}=\operatorname{sigmoid}\!\left(\frac{p-c_{j-1}}{\tau}\right),
    \qquad
    \widetilde m=m+s-\operatorname{sg}(s),
    \label{eq:appendix-selection-gradient}
\end{equation}
where $\operatorname{sg}$ stops gradients and $\tau=0.01$.
The mask is scattered back to the original spatial order before mixing.
Budget and binarization penalties are
\begin{equation}
    \mathcal L_{\mathrm{budget}}=(\langle s\rangle-0.375)^2,
    \qquad
    \mathcal L_{\mathrm{binary}}=\langle s(1-s)\rangle,
\end{equation}
where the averages cover the local minibatch and spatial positions.
Fixed-budget training uses a hard top-96 mask with a sigmoid surrogate
of normalized logits, temperature 0.7, and a log-odds bias for ratio 0.375.
The budget penalties regularize the soft masks during training; evaluation
uses hard selection.

\subsubsection{Loss}
\label{app:reconstruction-losses}

Alignment pretraining combines image $L_1$, LPIPS~\citep{zhang2018lpips},
feature $L_1$ between $\bar F_{\mathrm{1D}}$ and $F_{\mathrm{2D}}$, and
feature-moment matching, with weights $1$, $0.5$, $0.05$, and $0.05$,
respectively. Moment matching sums the $L_1$ differences of per-channel
spatial means and standard deviations.

For joint fixed-budget and top-$p$ training, the reconstruction objective
in Eq.~\eqref{eq:MixTok-loss} is supplemented with feature and routing
losses. Their coefficients are listed in
Table~\ref{tab:appendix-reconstruction-losses}. DINOv2~\citep{oquab2023dinov2}
ViT-S/14 and CLIP~\citep{radford2021clip} ViT-B/32 provide frozen,
normalized image features. We use $L_1$ distances, averaging DINOv2's
class-token and patch-token losses. Both backbones receive images resized
to $224\times224$ and normalized using their respective input statistics.
The discriminator uses a frozen ConvNeXt-Base~\citep{liu2022convnext}
backbone with trainable convolutional heads. Its feature-matching term
compares real and reconstructed features using $L_1$ distance.
Adversarial training uses a hinge discriminator loss and the negative
mean fake score for the generator, with LeCam
regularization~\citep{tseng2021lecam} of weight $10^{-3}$ and EMA decay 0.999.

\begin{table}[H]
    \centering
    \caption{Loss coefficients during joint reconstruction training.
    The image terms are implemented as $2(\mathcal L_1+0.3\mathcal L_{\mathrm{LPIPS}})$.
    Adversarial and discriminator feature-matching terms follow the
    activation schedule in Appendix~\ref{app:reconstruction-schedule}.}
    \label{tab:appendix-reconstruction-losses}
    \begin{tabular}{lc@{\qquad}lc}
        \toprule
        Term & Weight & Term & Weight \\
        \midrule
        Image $L_1$ & 2.0 & Scorer supervision & 1.0 \\
        LPIPS & 0.6 & Budget penalty & 20.0 \\
        DINOv2 feature loss & 0.5 & Binarization penalty & 0.01 \\
        CLIP feature loss & 0.5 & Generator adversarial loss & 0.12 \\
        Discriminator feature matching & 0.5 & & \\
        \bottomrule
    \end{tabular}
\end{table}

\subsection{Generation Training and Sampling}
\label{app:generation-training}

\subsubsection{Generation Grid Scorer}

The frozen reconstruction model supplies teacher logits
$\rho^{\mathrm{rec}}$, while the student predicts $\rho^{\mathrm{gen}}$
from $\bar F_{\mathrm{1D}}$ alone. The distillation loss is
\begin{equation}
    \mathcal L_{\mathrm{distill}}=
    D_{\mathrm{KL}}\!\left(q^{\mathrm{rec}}\,\|\,q^{\mathrm{gen}}\right)
    +\alpha\operatorname{SmoothL1}
    (\rho^{\mathrm{gen}}-\langle\rho^{\mathrm{gen}}\rangle,
     \rho^{\mathrm{rec}}-\langle\rho^{\mathrm{rec}}\rangle),
\end{equation}
where each distribution is a temperature-scaled softmax over locations,
and logit means are computed separately for each image.
We train the scorer for 20,000 updates on eight GPUs with batch size 32
per GPU (256 in total), temperatures of 2 for both distributions, and
$\alpha=0.05$. AdamW uses a constant learning rate of $2\times10^{-4}$,
weight decay $10^{-2}$, $(\beta_1,\beta_2)=(0.9,0.999)$, and gradient
clipping at norm 1. The frozen teacher forward pass uses BF16.
Distillation supervises the full spatial distribution and does not depend
on a top-$p$ cutoff; $p=0.75$ is used for subsequent location selection.

\subsubsection{Embedding Adapter and 2D Generator}

\paragraph{Cached training pairs.}
Generation training uses two cached views of each ImageNet training image:
an ADM-style center crop and its horizontal flip. Each view provides
paired 1D codes and original-image 2D targets. For adapter training, we also cache tokens obtained by re-encoding
the 1D base reconstructions with the frozen 2D tokenizer. 
A separate cache stores the locations selected by the frozen Generation
Grid Scorer with $T=2$ and $p=0.75$. The 2D generator batches examples
with equal $K$ to avoid padding the selected targets.

\paragraph{Embedding regression.}
Let $t_i^{\mathrm{base}}$ be the re-encoded base-image token and
$E_{\mathrm{tok}}^{\mathrm{pre}}$ the fixed pretrained Halton embedding
table. The adapter loss averages over every spatial location:
\begin{equation}
    \mathcal L_A=
    \frac{1}{Nd\,v_E}\sum_{i=1}^{N}
    \left\|A(\bar F_{\mathrm{1D}})_i-
    E_{\mathrm{tok}}^{\mathrm{pre}}(t_i^{\mathrm{base}})\right\|_2^2,
\end{equation}
where $d$ is the token embedding dimension and $v_E$ is the fixed
variance of all entries in the pretrained embedding table,
excluding the mask-token row. The tokenizer and
embedding targets remain frozen. Adapter training uses BF16, while base
features and the regression loss are computed in FP32.

\paragraph{Masked generator training.}
After adapter training, $A$ remains frozen and all Halton generator
parameters are fine-tuned. With equal probability per example, we mask
selected positions using either the official arccos random-mask schedule
or a Halton-prefix mask. The latter samples a completed-round index
uniformly from 0 to 31 and masks the unrevealed suffix of the selected
Halton order. Every example retains at least one masked target.
Cross-entropy is averaged only over masked selected positions, using
the original-image 2D codes as labels. The class condition is dropped
with probability 0.1. Adapter and generator training use AdamW with
$(\beta_1,\beta_2)=(0.9,0.999)$ and gradient clipping at norm 1.

\begin{table}[H]
    \centering
    \caption{Continuous-context generation training settings.
    Each model is trained on eight GPUs.}
    \label{tab:appendix-generation-training}
    \setlength{\tabcolsep}{7pt}
    \begin{tabular}{lcc}
        \toprule
        Setting & Embedding Adapter & Halton generator \\
        \midrule
        Optimizer updates & 20,000 & 20,000 \\
        Global batch size & 512 & 256 \\
        Learning rate & $5\times10^{-4}$ & $2\times10^{-5}$ \\
        Weight decay & 0.01 & 0.03 \\
        Warm-up updates & 500 & 2,500 \\
        Training precision & BF16 & BF16 \\
        \bottomrule
    \end{tabular}
\end{table}

The adapter uses a cosine learning-rate schedule over 20,000 updates.
The generator warms up for 2,500
updates, holds its learning rate until update 4,000, then decays by a
cosine schedule to $2\times10^{-6}$ at update 20,000.

\subsubsection{Sampling Configuration}

For the main generation results, the pretrained TiTok MaskGIT uses eight
rounds, guidance scale 4.5 with linear guidance scheduling, and
randomization temperature 9.5; softmax temperature annealing is disabled.
The Generation Grid Scorer selects locations with $T=2$ and $p=0.75$.
The Halton generator uses eight rounds with an arccos schedule,
guidance weight $w=3.0$, and categorical sampling at temperature 1.
Its classifier-free guidance combines logits as
$\ell_{\mathrm{cfg}}=(1+w)\ell_{\mathrm{cond}}-w\ell_{\mathrm{uncond}}$.
The mask-token logit is excluded from sampling, and the context
embeddings use scale 1. Thus, the final 1D/2D guidance settings are 4.5/3.0, with $8+8$ sampling rounds.

\subsection{Evaluation Details}
\label{app:evaluation}

\paragraph{Reconstruction.}
Validation uses the center-crop implementation supplied with LlamaGen and the
EMA tokenizer weights. The final allocation parameters are $T=1$ and
$p=0.994897$, with no additional minimum or maximum token clamp.
The threshold $p$ was calibrated on a randomly sampled subset of the ImageNet
training set after the 20-epoch fixed-budget stage to match the target average
budget of approximately 96 2D tokens.
These evaluation settings differ from the top-$p$ fine-tuning values
$T=2$ and $p=0.91$ in Table~\ref{tab:appendix-reconstruction-training}.
For rFID, original and reconstructed images are clipped to $[0,1]$,
scaled by 255, rounded to uint8, and evaluated with 2,048-dimensional
Inception features through TorchMetrics. PSNR is computed for each image
in $[0,1]$ before averaging over the validation set; it is not computed
from a single dataset-averaged MSE.

\paragraph{Reconstruction baseline configurations.}
\label{app:reconstruction-baselines}
For Table~\ref{tab:reconstruction-baselines}, TiTok BL-128 uses the
updated single-stage VQ training recipe. One-D-Piece uses the first
128 tokens of the released L-256 checkpoint. FlexTok uses discrete
FSQ~\citep{mentzer2024fsq} tokens with a flow decoder, evaluated with
25 decoding steps, guidance scale 7.5, adaptive projected guidance,
and seed 0.

\paragraph{Generation.}
The ADM evaluator~\citep{dhariwal2021adm} uses the official
\texttt{VIRTUAL\_imagenet256\_labeled.npz} reference statistics.
For the main results, we use sampling seeds 1--4, each generating
50,000 images with 50 examples per class. We report the mean and sample
standard deviation of the four independently computed gFID values
(Table~\ref{tab:generation-seeds}). These repeats measure sampling
variability for a fixed checkpoint and sampling configuration.
The reported token count averages $L+K(x)$ over all four runs;
position indices, class conditioning, and internal context embeddings
are excluded from the content-token count.

\begin{table}[H]
    \centering
    \caption{Generation quality before and after sparse 2D refinement
    across sampling seeds. Each seed uses 50,000 images. The final row
    reports the mean $\pm$ sample standard deviation across four runs.
    Token statistics refer to the final representation.}
    \label{tab:generation-seeds}
    \setlength{\tabcolsep}{3.5pt}
    \begin{tabular}{lcccc}
        \toprule
        Seed & \shortstack{1D base\\gFID $\downarrow$}
        & \shortstack{Full refinement\\gFID $\downarrow$}
        & Mean $K$ & \shortstack{Mean total\\tokens} \\
        \midrule
        1 & 4.2758 & 1.9518 & 98.66 & 130.66 \\
        2 & 4.3697 & 1.9509 & 98.53 & 130.53 \\
        3 & 4.3611 & 1.9597 & 98.75 & 130.75 \\
        4 & 4.3318 & 1.9420 & 98.69 & 130.69 \\
        \midrule
        Mean $\pm$ SD & $4.3346\pm0.0424$ & $1.9511\pm0.0072$
        & $98.66\pm0.09$ & $130.66\pm0.09$ \\
        \bottomrule
    \end{tabular}
\end{table}

The 1D base uses 32 generated tokens and eight sampling rounds,
decoded through MixTok's spatial expansion, feature alignment, and
shared decoder. Starting from the same generated 1D tokens, sparse
2D refinement improves mean gFID from 4.3346 to 1.9511 with eight
additional sampling rounds and approximately 131 total content tokens.

\subsection{Distribution of 2D Token Counts}
\label{app:generation-k-distribution}

\begin{figure}[H]
    \centering
    \includegraphics[width=0.95\linewidth]{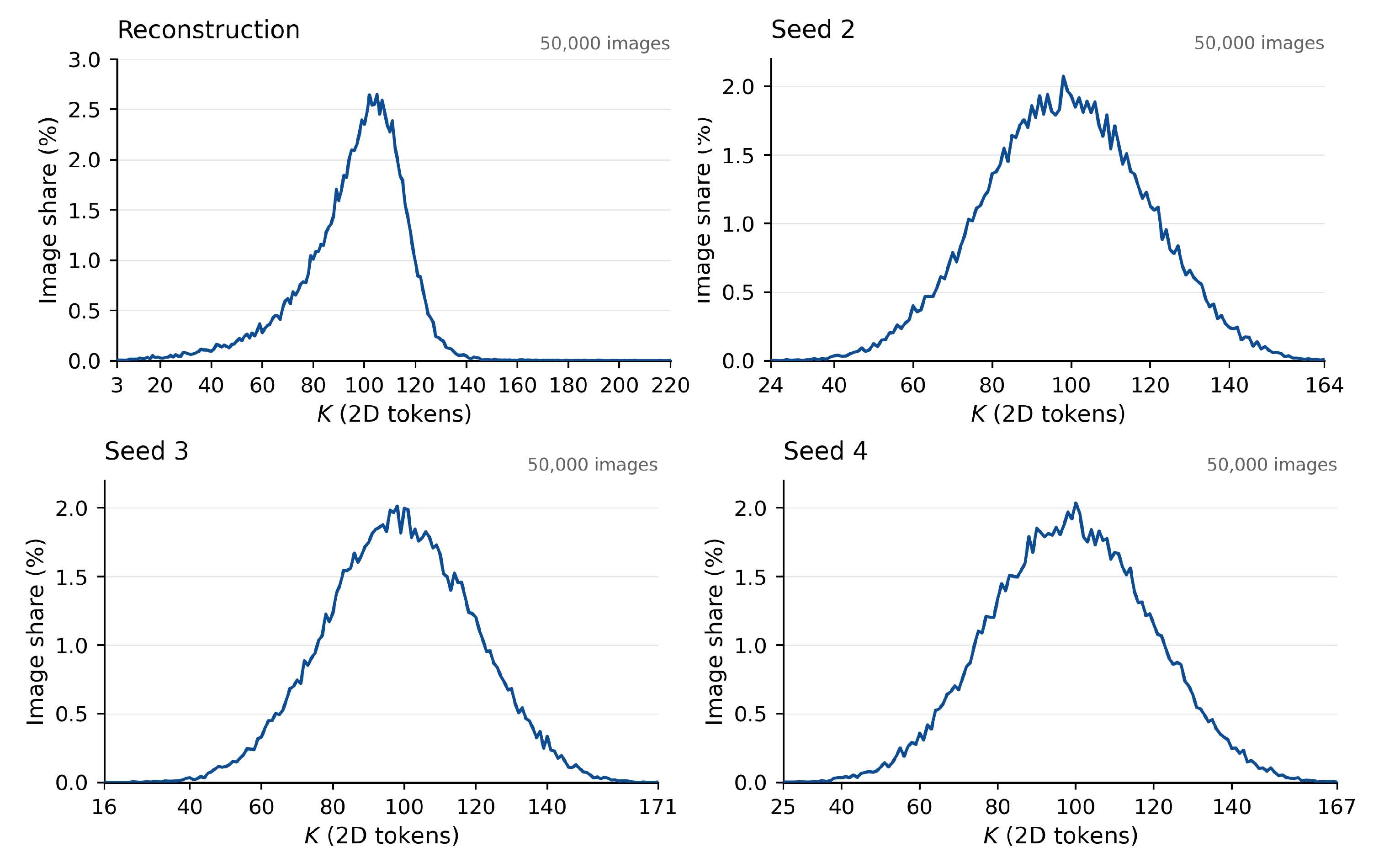}
    \caption{Distribution of selected 2D token counts $K$ for
reconstruction on the ImageNet validation set and generation with
sampling seeds 2--4, with 50,000 images per panel. The vertical
axis shows the percentage of images assigned each integer $K$.
The fixed 32 1D tokens are excluded. Each panel spans the smallest
to largest observed $K$, retaining zero-count bins within this range.}
    \label{fig:generation-k-distribution}
\end{figure}

Figure~\ref{fig:generation-k-distribution} shows image-dependent
refinement budgets in both reconstruction and generation.
For generation seeds 2--4, mean $K$ ranges from 98.53 to 98.75,
with similar distributions across seeds.

\subsection{Bitrate Accounting}
\label{app:bitrate-accounting}

Table~\ref{tab:appendix-bitrate} supplements the reconstruction results
in Table~\ref{tab:reconstruction-baselines} with estimated bitrates
under fixed-width coding, using the same checkpoints and decoding
settings. For MixTok, 12 / 14 denotes the bit widths of the 1D / 2D
token indices. Each token index from a vocabulary of size
$V$ is assigned $\lceil\log_2 V\rceil$ bits. The baseline configurations
use fixed evaluation lengths and predefined token orders, requiring
no per-image position or length fields. FlexTok's FSQ levels
$[8,8,8,5,5,5]$ yield 64,000 possible codes, requiring 16 bits per
packed token index.

For MixTok, we account for the selected locations using a 256-bit
binary mask over the $16\times16$ grid, with the selected 2D token
indices stored in raster order. The mask determines both the positions
and $K(x)$, so no separate length field is needed. With 12 bits per
1D token and 14 bits per 2D token, the average bitrate is
\begin{equation}
    \overline{R}_{\mathrm{MixTok}}
    = \frac{32\times12 + 14\,\mathbb{E}[K(x)] + 256}{256\times256}.
\end{equation}
The measured mean $\mathbb{E}[K(x)]=96.1093$ gives 0.03030 bpp,
including 0.00390625 bpp for spatial side information.
These are calculated representation payloads; no entropy coding is
assumed, and file headers, byte padding, and shared model weights are
excluded. Equal content-token counts can therefore correspond to
different bitrates.

\begin{table}[H]
    \centering
    \caption{Reconstruction quality and estimated fixed-width bitrate
    on ImageNet $256\times256$. Total bpp includes spatial side
    information. Token counts are rounded for display; bitrate uses
    unrounded counts.}
    \label{tab:appendix-bitrate}
    \small
    \setlength{\tabcolsep}{3pt}
    \renewcommand{\arraystretch}{1.15}
    \begin{tabularx}{\linewidth}{@{}>{\raggedright\arraybackslash}Xrrrrr@{}}
        \toprule
        Tokenizer & \shortstack{Avg. content\\tokens}
        & \shortstack{Bits per\\token index}
        & \shortstack{Total bpp\\$\downarrow$}
        & rFID $\downarrow$ & \shortstack{PSNR\\(dB) $\uparrow$} \\
        \midrule
        TiTok S-128~\citep{yu2024titok}
        & 128 & 12 & 0.02344 & 1.7222 & 17.72 \\
        TiTok BL-128 VQ~\citep{kim2025tatitok}
        & 128 & 13 & 0.02539 & 1.5063 & 19.83 \\
        One-D-Piece L-256~\citep{miwa2025onedpiece}
        & 128 & 12 & 0.02344 & 1.4406 & 17.71 \\
        FlexTok d18-d28~\citep{bachmann2025flextok}
        & 128 & 16 & 0.03125 & 1.1422 & 16.76 \\
        DetailFlow-16~\citep{liu2025detailflow}
        & 128 & 13 & 0.02539 & 1.1370 & 19.46 \\
        \midrule
        LlamaGen VQ-16~\citep{sun2024llamagen}
        & 256 & 14 & 0.05469 & 2.1872 & 20.78 \\
        MaskGIT-VQGAN~\citep{chang2022maskgit}
        & 256 & 10 & 0.03906 & 2.1304 & 18.15 \\
        \midrule
        \textbf{MixTok (ours)}
        & 128.11 & 12 / 14 & 0.03030 & 1.1950 & 19.51 \\
        \bottomrule
    \end{tabularx}
\end{table}

\subsection{Generation Sampling-Round Ablation}
\label{app:stage2-steps}

Using the main generation checkpoint, we vary only the 2D sampling rounds.
The 1D stage uses eight rounds, and all other sampling settings are held
fixed. Each setting generates 50,000 images with sampling seed 4.
These single-seed results are reported separately from the four-seed
main results in Table~\ref{tab:generation-seeds}.

\begin{table}[H]
    \centering
    \caption{Effect of 2D sampling rounds with all other settings fixed.
    All settings use sampling seed 4.}
    \label{tab:stage2-steps}
    \begin{tabular}{ccc}
        \toprule
        2D rounds & Total rounds & gFID $\downarrow$ \\
        \midrule
        4 & 12 & 2.0314 \\
        8 & 16 & 1.9420 \\
        16 & 24 & 1.9204 \\
        32 & 40 & 1.9191 \\
        48 & 56 & 1.9242 \\
        64 & 72 & 1.9355 \\
        \bottomrule
    \end{tabular}
\end{table}

Increasing the 2D stage from 4 to 8 rounds substantially improves gFID, while further increases yield diminishing returns, with performance saturating around 24--40 total rounds. We therefore use 8+8 rounds as the main efficiency-oriented configuration.

\subsection{Generation Efficiency}
\label{app:generation-efficiency}

\paragraph{Generation table configurations.}
In Table~\ref{tab:generation-baselines}, steps count sampling rounds;
DetailFlow-32 uses eight causal plus 31 group-refinement passes.
The starred MaskGIT result is officially reported for the released
checkpoint and is not reevaluated with our ADM protocol.
Halton-MaskGIT-L uses its native $384\times384$ output and a
$24\times24$ token grid; the other models output $256\times256$.
MixTok reports mean gFID over seeds 1--4 (sample SD: 0.0072), with
50,000 images per seed, as detailed in Appendix~\ref{app:evaluation}.

\paragraph{Latency.}
We measure each model on one RTX 5090 with batch size 1 and ImageNet
class 207. After 10 warm-up runs, we time 50 runs with sampling seeds
4--53 and report the median. Sampling-only and end-to-end runs are paired
by seed with alternating order; Table~\ref{tab:generation-baselines}
reports end-to-end latency. We synchronize CUDA before and after each
wall-clock measurement. Timing covers class-conditioned generation
through final RGB decoding, with outputs remaining on the GPU; model
loading, file I/O, and FID evaluation are excluded. For MixTok, this
includes both generators, spatial expansion and feature alignment,
the Generation Grid Scorer, and the Embedding Adapter.

\paragraph{Implementation settings.}
We use PyTorch 2.10.0 with CUDA 12.8, four CPU threads per process,
TF32 enabled, and cuDNN benchmarking. We retain each implementation's
attention kernels, caches, guidance, and sampling loops, without
\texttt{torch.compile} or CUDA Graphs. BF16 autocast is enabled while
preserving explicit precision settings: MixTok's latent decoder,
scorer, and RGB decoder use FP32, and AdaGen's backbone uses internal
FP16 autocast. All models are benchmarked sequentially on the same RTX 5090,
each in a separate process, on a shared host with unlocked GPU clocks. These measurements
describe the tested implementations and workload; gFID values come
from the separate evaluations described in Appendix~\ref{app:evaluation}.

\paragraph{FLOPs and parameters.}
We count one complete class-207, seed-4 inference with PyTorch's
\texttt{FlopCounterMode}, counting a multiply-add as two FLOPs.
The counts include matrix multiplications, convolutions, supported
attention operations, all sampling rounds, classifier-free guidance,
and final image decoding. Elementwise operations, sorting, sampling,
and data movement are excluded from FLOPs but included in latency.
FLOPs are reported per image and are not averaged over the 50,000-image
quality evaluations. Parameter counts for measured models include all
parameters in the instantiated generators and auxiliary generation
modules, excluding tokenizers. AdaGen includes its policy network;
MixTok includes both generators, its scorer, and adapter. TiTok-S-128
and One-D-Piece-L-256 contain 294.991M and 177.374M parameters under this
counting convention, respectively. Halton-L uses its native
$384\times384$ output and is not a matched-resolution compute comparison.